\documentclass{article}

\usepackage{microtype}
\usepackage{graphicx}
\usepackage{subcaption}
\usepackage{booktabs} % for professional tables
\usepackage{placeins}
\usepackage{float}
\usepackage{hyperref}

\usepackage[accepted]{icml2026}

\usepackage{amsmath}
\usepackage{amssymb}
\usepackage{mathtools}
\usepackage{amsthm}

\usepackage[capitalize,noabbrev]{cleveref}

\theoremstyle{plain}

\theoremstyle{definition}

\theoremstyle{remark}

\usepackage[textsize=tiny]{todonotes}

\icmltitlerunning{\icmltitlerunning{Source-Grounded Synthetic Notes from Structured EHR}}

\begin{document}

\twocolumn[
  \icmltitle{A Multi-Agent Pipeline for Source-Grounded Synthetic Note Generation from Longitudinal Structured EHR}



\icmlsetsymbol{equal}{*}

  \begin{icmlauthorlist}
    \icmlauthor{Nina Fatehi}{comp}
    \icmlauthor{Reihaneh Hassanzadeh}{comp}
    \icmlauthor{Meysam Ghaffari}{comp}
    \icmlauthor{Animesh Agrawal}{comp}
    \icmlauthor{Carlos Morato}{comp}
  \end{icmlauthorlist}
  \icmlaffiliation{comp}{Optum AI, UnitedHealth Group, Minneapolis, Minnesota, USA}
  \icmlcorrespondingauthor{Carlos Morato}{carlos.morato@optum.com}
  \icmlkeywords{Machine Learning, ICML}
  \vskip 0.15in
]
\printAffiliationsAndNotice{}  % no special 
\begin{abstract}
Structured EHR is abundant but sparse, coded, and difficult to use directly for note-centric clinical modeling. We present MedNotes, a multi-agent synthetic data generation pipeline that converts longitudinal structured EHR into source-grounded clinical note representations under explicit quality control. MedNotes  treats structured-data-to-text synthesis as a closed-loop agentic process: a generator proposes a note, evaluator agents identify factual, coverage, structural, and hallucination-related failures, and an automatic routing component accepts, revises, or rejects the draft. On 1{,}485 EHRSHOT encounters, MedNotes achieves a 91.4\% pass rate, with mean factual accuracy of 0.980, completeness of 99.1\%, structural fidelity of 0.761, and 0.028 critical hallucinations per encounter. Iterative refinement improves acceptance from 69.4\% to 91.4\%. The resulting synthetic corpus improves downstream CPT prediction and paragraph-level section prediction when combined with limited real data.
\end{abstract}
\vspace{-0.08in}
\section{Introduction}
Structured EHR is abundant but difficult to use directly for many downstream clinical
language-modeling tasks because it is sparse, coded, longitudinal, and often lacks
shareable free-text notes. At the same time, authentic clinical notes are valuable for
model development but difficult to release at scale because of privacy, governance, and
annotation constraints. This has motivated growing interest in synthetic clinical text and
source-grounded data generation, especially when synthetic artifacts preserve task-relevant
structure while remaining traceable to the underlying data
\citep{nadas2025synthetic,lupidi2024source2synth,amad2025improving}. 
Recent work on tabular generation and structured-data serialization suggests that language-model-style representations can serve as useful interfaces for structured records 
\citep{borisov2023language,hegselmann2023tabllm,solatorio2023realtabformer,hollmann2023tabpfn}.
Most clinical note generation work assumes richer source modalities, such as encounter
dialogue, audio, or paired clinician-authored notes, as in MEDIQA-Chat and ACI-Bench
\citep{abacha2023overview,yim2023aci}. In many realistic settings, however, the available
data are longitudinal structured EHR: diagnoses, medications, procedures, laboratory
results, vital signs, and coded observations. EHRSHOT is a representative example of this
structured-only setting \citep{wornow2023ehrshot}. This creates a modality gap: many
downstream note-centric models expect text, while the most accessible clinical data source
is often coded EHR.

We study this setting as a quality-gated synthetic data generation problem. The goal is
not to reproduce clinician-authored documentation, but to construct source-grounded
clinical note representations that can support downstream adaptation, benchmarking, and
structured-data learning. We instantiate the target format as SOAP because its sectioned
anatomy provides a useful interface for controlled generation: it separates evidence-bearing
content from interpretation and planning, supports section-aware feedback, and enables
structural consistency checks.

We introduce MedNotes, a multi-agent pipeline for converting longitudinal structured EHR
into source-grounded synthetic clinical notes. MedNotes treats structured-data-to-text
synthesis as a closed-loop agentic process: a generator proposes a note, evaluator agents
identify factual, coverage, structural, and hallucination-related failures, an aggregator
constructs repair signals and preservation anchors, and a router accepts, revises, or rejects
the draft under explicit quality constraints. This adapts iterative feedback and
textual-gradient ideas \citep{madaan2023selfrefine,shinn2023reflexion,pryzant2023proteger}
to quality-gated synthetic corpus construction from structured clinical data.
\section{Methods and Cohort}
\label{sec:methods}

We present MedNotes, a controlled synthetic data generation pipeline that transforms
longitudinal structured EHR into source-grounded clinical note representations under
explicit quality gating. We instantiate the target representation as SOAP because its
sectioned structure supports modular prompting, feedback assignment, and structural
evaluation, but the method is intended more broadly as quality-gated structured-data-to-text
synthesis.

\subsection{Task Formulation}
\label{sec:task_formulation}

Let $e_i$ denote the current encounter for patient $p$ at index $i$, and let $H_i^{(k)}$
denote up to $k=2$ prior encounters used as longitudinal context. The generator input is
\begin{equation}
\label{eq:input}
x_i = \left(e_i, H_i^{(k)}\right),
\end{equation}
where encounters contain structured clinical information such as diagnoses, medications,
procedures, laboratories, and vital signs after code resolution into human-readable text.
Given $x_i$, the system generates a synthetic sectioned clinical note $n_i$.

Accepted outputs form a synthetic corpus
\begin{equation}
\label{eq:synthetic_corpus}
\mathcal{D}_{\mathrm{syn}}=\{(x_i,n_i): \mathrm{PASS}(n_i,x_i)=1\},
\end{equation}
intended for downstream adaptation, benchmarking, and representation learning when
authentic notes or paired supervision are unavailable. Because structured EHR rarely
contains recoverable patient-reported narrative, the Subjective section is treated
conservatively: the model may summarize supported contextual information but must abstain
from inventing unsupported patient-reported details.

For each candidate note, MedNotes evaluates four quality dimensions:
\begin{equation}
\label{eq:synthetic_corpus}
\begin{aligned}
f_i &= \mathrm{Fact}(n_i,x_i) \in [0,1],\\
c_i &= \mathrm{Comp}(n_i,x_i) \in [0,100],\\
s_i &= \mathrm{SFS}(n_i,x_i) \in [0,1],\\
h_i^{\mathrm{crit}} &= \mathrm{CritHall}(n_i,x_i) \in \mathbb{N}_0,
\end{aligned}
\end{equation}
where $f_i$ is factual accuracy, $c_i$ is entity completeness, $s_i$ is structural fidelity Score (SFS)
under the target note format, and $h_i^{\mathrm{crit}}$ is the number of critical
hallucinations. A note is accepted only if
% \begin{equation}
% \label{eq:pass}
% \mathrm{PASS}(n_i,x_i)=
% \mathbb{I}\left[
% f_i \geq 0.95 \wedge c_i \geq 90 \wedge s_i \geq 0.67 \wedge h_i^{\mathrm{crit}}=0
% \right].
% \end{equation}
\begin{equation}
\label{eq:pass}
\begin{aligned}
\mathrm{PASS}(n_i,x_i)
= \mathbb{I}\big[&
f_i \ge 0.95 \;\wedge\; c_i \ge 90 \\
&\wedge\; s_i \ge 0.67
\;\wedge\; h_i^{\mathrm{crit}} = 0
\big].
\end{aligned}
\end{equation}
Generation is iterative. Let $n_i^{(0)}$ be the initial draft. After round $r$, evaluator
agents produce section-specific corrective feedback $\Gamma_i^{(r)}$ and positive anchors
$A_i^{(r)}$ for already-correct content:
\begin{equation}
\label{eq:refinement}
\begin{aligned}
n_i^{(0)} &= G(x_i),\\
n_i^{(r+1)} &= G\!\left(x_i,n_i^{(r)},\Gamma_i^{(r)},A_i^{(r)}\right).
\end{aligned}
\end{equation}
The pipeline runs for at most $R_{\max}=3$ rounds. If no draft satisfies the acceptance
criterion, the system abstains and returns no note.

\subsection{Dataset and Cohort}
\label{sec:dataset}

We evaluate on 100 patients from EHRSHOT, a de-identified longitudinal EHR benchmark
spanning 2009--2022 \citep{wornow2023ehrshot}. Patients were selected by balanced
stratified sampling across diabetes, hypertension, obesity, and a general cohort. The final
evaluation set contains 1{,}485 encounters, with 10--20 encounters per patient. Each
encounter includes structured codes from multiple vocabularies, including ICD-10-CM,
RxNorm, CPT, LOINC, and SNOMED. All codes are resolved to human-readable descriptions
before generation to reduce ambiguity and code-interpretation errors.

\subsection{MedNotes Pipeline}
\label{sec:pipeline}

\begin{figure}[t]
    \centering
    \includegraphics[width=\linewidth]{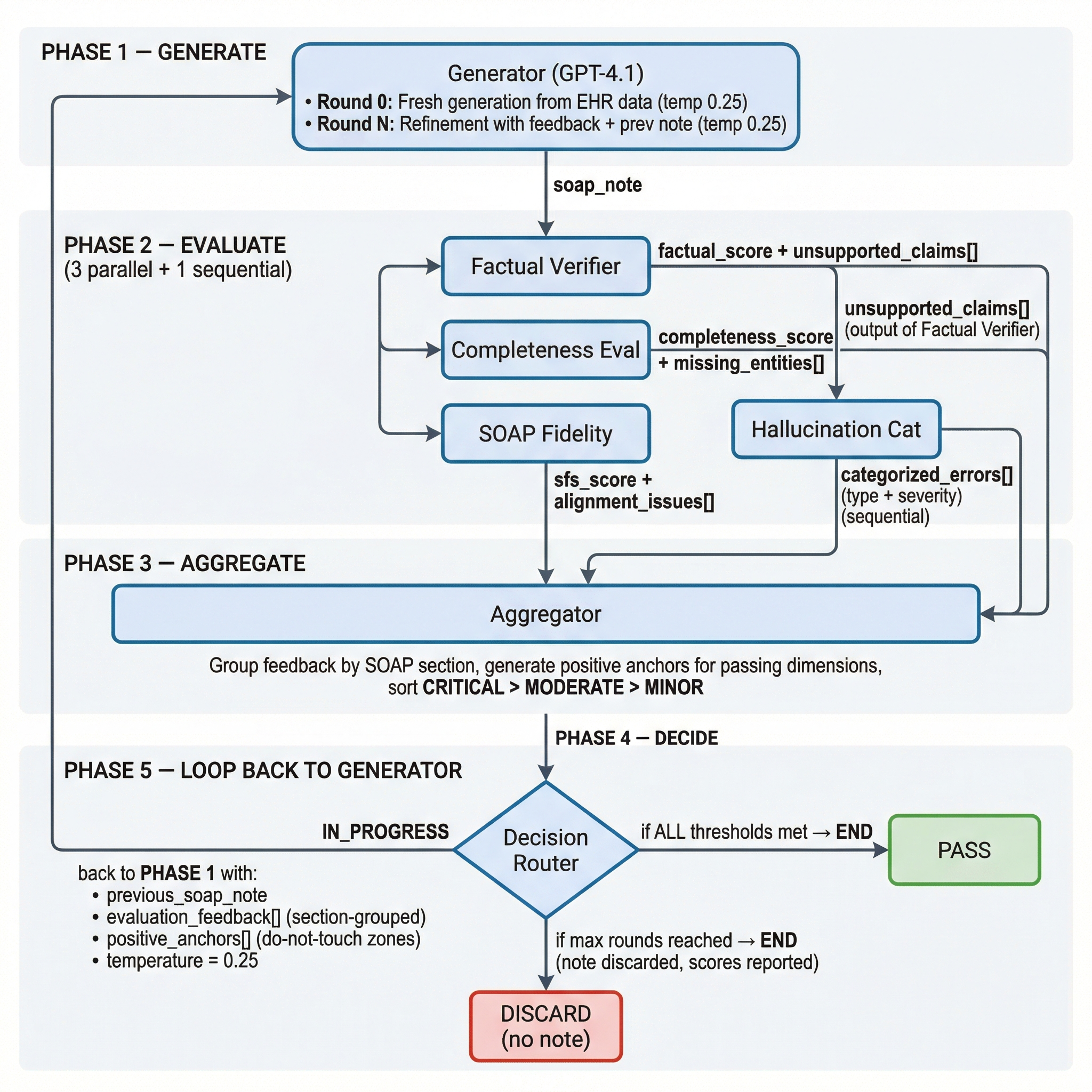}
    \caption{MedNotes pipeline architecture. Each encounter passes through generation,
    evaluation, aggregation, and decision stages. Factual verification, entity completeness,
    and structural fidelity are computed in parallel, followed by hallucination
    categorization. Evaluator outputs are converted into section-specific repair signals and
    positive anchors for the next round.}
    \label{fig:mednotes-pipeline}
\end{figure}

MedNotes treats structured-EHR-to-text synthesis as a closed-loop agentic control problem
rather than a one-shot prompting task. The generator proposes an initial note from the
current encounter and limited longitudinal context. Evaluator agents then decompose note
quality into complementary dimensions: factual support against the source EHR, recall-style
coverage of structured entities, structural consistency across note sections, and severity of
unsupported claims. The aggregator compiles evaluator outputs into section-specific repair
signals and positive anchors, and the automatic routing component either accepts the note, sends it back
for revision, or rejects the encounter after the round budget is exhausted.

The generator is GPT-4.1 (Azure OpenAI; temperature 0.25) with a conservative rule-based
system prompt. In the first round, it generates a note from the current encounter and up to
two prior encounters. In later rounds, it is re-invoked with the previous note,
section-grouped feedback, and positive anchors. The prompt emphasizes resolved clinical
descriptions rather than raw codes, comprehensive use of current-encounter evidence,
qualified historical references, conservative handling of missing patient-reported content,
and abstention from unsupported inference.

The evaluation layer uses specialized agents to measure source-grounded quality. A factual consistency evaluator compares note claims against structured EHR data.; a completeness evaluator measures
coverage of clinically significant entities; a structural fidelity evaluator scores
cross-section consistency; and a hallucination categorizer identifies and categorizes  unsupported claims by
type and severity. These outputs are merged into targeted feedback for the next generation
round. A note is retained in $\mathcal{D}_{\mathrm{syn}}$ only when all thresholds are
satisfied jointly; otherwise, the system revises the note or abstains after three rounds.
\section{Experiments and Results}
\label{sec:experiments_results}
We evaluate three questions. First, can MedNotes construct source-grounded synthetic notes
from structured longitudinal EHR under explicit quality gating? Second, how much does
iterative evaluator-guided refinement improve acceptance? Third, do the accepted synthetic
notes carry useful signal for downstream note-centric prediction tasks? The primary quality
metric is encounter-level acceptance under Eq.~\ref{eq:pass}; we also report final-round
factual accuracy, entity completeness, structural fidelity, critical hallucinations, and total
categorized error burden. Supporting development studies, including prompt optimization,
lookback-window sensitivity, relevance-based history selection, the full error-recovery
breakdown, and further downstream task details, are provided in the appendix.
\subsection{Quality-Gated Synthesis on EHRSHOT}
\label{sec:overall_results}
Table~\ref{tab:overall} summarizes aggregate system performance. MedNotes achieves a
91.4\% encounter-level pass rate, with 1{,}357 accepted encounters and 128 discarded
encounters. No API failures occurred during evaluation. Across final outputs, the system
achieves a mean factual accuracy of 0.980, mean completeness of 99.1\%, mean structural
fidelity of 0.761, and 0.028 critical hallucinations per encounter. These results indicate
that the multi-agent control loop can satisfy strict source-grounded quality criteria for most
encounters while preserving an explicit abstention mechanism for cases that remain too sparse or ambiguous. In a runtime benchmark over the same 1{,}485 encounters, MedNotes processed encounters in
76.8 seconds on average at an estimated API cost of \$0.0958 per encounter
(Appendix~\ref{app:cost_latency}); this reflects the deployed API configuration for this run,
and model versions and pricing may differ across configurations.
\begin{table}[h]
\centering
\caption{Overall results on the 100-patient evaluation cohort.}
\label{tab:overall}
\small
\setlength{\tabcolsep}{4pt}
\begin{tabular}{p{0.62\columnwidth} p{0.28\columnwidth}}
\hline
Metric & Value \\
\hline
Total encounters & 1,485 \\
Pass & 1,357 (91.4\%) \\
Discarded & 128 (8.6\%) \\
API failures & 0 (0.0\%) \\
Mean factual accuracy (final) & 0.980 \\
Mean completeness (final) & 99.1\% \\
Mean SFS (final) & 0.761 \\
Mean critical hallucinations / encounter & 0.028 \\
Mean total categorized error burden / encounter & 0.344 \\
\hline
\end{tabular}
\end{table}
\subsection{Reflection and Error Recovery}
\label{sec:reflection_results}
To isolate the contribution of iterative reflection, we analyzed the round at which each
encounter first satisfied the joint acceptance criterion. As shown in Table~\ref{tab:reflection},
1{,}031 encounters passed in Round~1, corresponding to 69.4\% of all encounters and
76.0\% of accepted encounters. Round~2 added 266 accepted encounters and Round~3 added
60 more, yielding a final pass rate of 91.4\%.

Without reflection, MedNotes would achieve only a 69.4\% pass rate. The iterative repair
loop therefore contributes a net gain of 22.0 percentage points. Among the 454 encounters
that failed in Round~1, reflection recovered 326, corresponding to a 71.8\% recovery rate.
A finer-grained analysis showed that isolated hallucination and factuality failures were more
recoverable than persistent structural or completeness failures. Full recovery breakdowns are in Appendix~\ref{app:error_analysis}.

\begin{table}[H]
\centering
\small
\caption{Pass rate by reflection round.}
\label{tab:reflection}
\begin{tabular}{lrrr}
\toprule
Round & Pass & \% of accepted & Cumulative \\
\midrule
Round 1 & 1{,}031 & 76.0\% & 69.4\% \\
Round 2 & 266 & 19.6\% & 87.3\% \\
Round 3 & 60 & 4.4\% & 91.4\% \\
Discarded & 128 & --- & --- \\
\bottomrule
\end{tabular}
\end{table}

\subsection{Preliminary Clinician Review}
\label{sec:clinician_review}

We obtained exploratory feedback from a practicing physician on a small subset of generated
notes, focusing on source support, clinically important omissions, and overall usefulness.
This was intended as a qualitative expert assessment rather than a formal human-evaluation
study.

Across reviewed cases, the physician feedback was consistent with the intended conservative operating point of MedNotes. Accepted notes were  readable and source-grounded, but often reflected the sparsity of the underlying structured EHR: they summarized coded events and available context rather than adding unsupported narrative detail, exam findings, or clinical reasoning absent from the source. This supports our framing of the current pass criterion as a source-grounded quality gate for synthetic corpus construction, rather than a guarantee of clinician-level documentation utility.

\subsection{Downstream Task Utility}
\label{sec:downstream_results}

We next tested whether the accepted synthetic corpus carries reusable note-centric signal in
downstream prediction tasks. We evaluate two settings: CPT prediction from notes and
paragraph-level section prediction.

\subsubsection{CPT Prediction from Synthetic Notes}
\label{sec:cpt_downstream_results}

Given a clinical note, the model predicts the set of CPT codes associated with the encounter.
This is formulated as a multi-label prediction problem. We compare an untuned Llama-3 8B
baseline, fine-tuning on 100 MIMIC note--CPT pairs, and a mixed-data setting using synthetic
notes plus 100 MIMIC examples. Table~\ref{tab:cpt_downstream_results} reports performance
on a held-out MIMIC test set~\citep{johnson2016mimiciii}. The mixed synthetic+MIMIC setting achieved the best
performance across all reported metrics, suggesting that synthetic notes provide useful
task-relevant supervision when combined with limited real data.
\begin{table}[th]
\centering
\caption{CPT prediction results on the held-out MIMIC test set.}
\label{tab:cpt_downstream_results}
\scriptsize
\setlength{\tabcolsep}{3pt}
\begin{tabular}{p{0.40\columnwidth} l c c c}
\hline
Model & Avg. & P & R & F1 \\
\hline
Base Llama-3 8B Instruct & Macro & 0.0152 & 0.0634 & 0.0234 \\
 & Micro & 0.0366 & 0.0688 & 0.0478 \\
\hline
+ Fine-tuned on 100 MIMIC & Macro & 0.0173 & 0.0665 & 0.0244 \\
 & Micro & 0.0441 & 0.0516 & 0.0476 \\
\hline
+ Synthetic + 100 MIMIC & Macro & 0.2137 & 0.1034 & 0.1030 \\
 & Micro & 0.0827 & 0.1204 & 0.0980 \\
\hline
\end{tabular}
\end{table}
\subsubsection{Paragraph-Level Section Prediction}
\label{sec:section_prediction_results}

We also evaluate paragraph-level section prediction, where each paragraph is classified as
Subjective, Objective, Assessment, or Plan. Using MediSOAP as the evaluation benchmark~\citep{medisoap_github},
we compare training on 100 real notes against training on synthetic data combined with
100 real notes. As shown in Table~\ref{tab:section_prediction}, the combined model reaches
0.972 accuracy and 0.972 macro-F1, achieving higher results on 100 real notes
alone (see Appendix~\ref{app:section_prediction} for section-wise performance). This indicates that the synthetic corpus provides useful complementary supervision
for structural note understanding.
\vspace{0.20in}
\begin{center}
\small
\refstepcounter{table}
\label{tab:section_prediction}
\textit{Table~\thetable. Overall performance on the MediSOAP test set.}

\vspace{2pt}
\setlength{\tabcolsep}{4pt}
\begin{tabular}{@{}p{0.58\columnwidth}cc@{}}
\toprule
Model & Accuracy & Macro-F1 \\
\midrule
BERT fine-tuned on 100 MediSOAP notes & 0.785 & 0.752 \\
BERT fine-tuned on synthetic + 100 MediSOAP notes & \textbf{0.972} & \textbf{0.972} \\
\bottomrule
\end{tabular}
\end{center}

% \section{Discussion}
\section{Discussion}
\label{sec:discussion}
MedNotes is a quality-gated synthetic data generation framework for converting sparse,
coded, longitudinal EHR into source-grounded text representations. Rather than
introducing a new foundation model, its main contribution is a closed-loop agentic controller
for structured-data-to-text synthesis: evaluator agents identify unsupported claims, missing
entities, structural inconsistencies, and hallucination risks; the aggregator converts these
signals into repair instructions and preservation anchors; and the automatic routing component accepts, revises,
or discards drafts. This makes synthetic corpus construction more controlled than one-shot
prompting when generated text may later be used for downstream model training.

The results suggest that accepted synthetic notes can serve as useful intermediate
representations when authentic notes are unavailable, especially when combined with limited
real data. At the same time, preliminary clinician review clarifies the operating point:
source-groundedness and structural consistency do not guarantee clinical richness, particularly
when structured EHR lacks detailed subjective narrative, exam findings, or clinical reasoning.
Future work should improve salience-aware history selection and synthesis while preserving
the evidence constraints that make the corpus traceable.
\section{Conclusion}
\label{sec:conclusion}
We presented MedNotes, a multi-agent pipeline for quality-gated synthetic note generation from longitudinal structured EHR. The system couples generation, evaluator-guided refinement, and abstention to construct source-grounded clinical text representations from sparse coded records. On EHRSHOT, MedNotes achieves high acceptance under strict quality thresholds, and downstream experiments show that the resulting synthetic corpus provides useful signal for clinical code prediction and section prediction when paired with limited real data. Taken together these findings may support quality-gated synthetic notes as a practical bridge between structured EHR and note-centric machine learning, while motivating future work on richer clinical salience and context selection.
\vspace{0.08in}
\nocite{langley00}

\bibliography{example_paper}
\bibliographystyle{icml2026}

%%%%%%%%%%%%%%%%%%%%%%%%%%%%%%%%%%%%%%%%%%%%%%%%%%%%%%%%%%%%%%%%%%%%%%%%%%%%%%%
%%%%%%%%%%%%%%%%%%%%%%%%%%%%%%%%%%%%%%%%%%%%%%%%%%%%%%%%%%%%%%%%%%%%%%%%%%%%%%%
% APPENDIX
%%%%%%%%%%%%%%%%%%%%%%%%%%%%%%%%%%%%%%%%%%%%%%%%%%%%%%%%%%%%%%%%%%%%%%%%%%%%%%%
%%%%%%%%%%%%%%%%%%%%%%%%%%%%%%%%%%%%%%%%%%%%%%%%%%%%%%%%%%%%%%%%%%%%%%%%%%%%%%%
\newpage
% \appendix
% \onecolumn

% You can have as much text here as you want. The main body must be at most $8$
% pages long. For the final version, one more page can be added. If you want, you
% can use an appendix like this one.

% The $\mathtt{\backslash onecolumn}$ command above can be kept in place if you
% prefer a one-column appendix, or can be removed if you prefer a two-column
% appendix.  Apart from this possible change, the style (font size, spacing,
% margins, page numbering, etc.) should be kept the same as the main body.
\appendix
\onecolumn
\section{Cost and Latency Analysis}
\label{app:cost_latency}

We measured runtime and API cost over the 1{,}485-encounter evaluation run. Cost estimates
use recorded token usage from the deployed API configuration and the model
prices at the time of execution. These costs exclude engineering, storage, orchestration
overhead, and human review time.

\begin{table}[h]
\centering
\small
\caption{System cost and latency analysis over 1{,}485 encounters. Token counts are empirical estimates and may vary with encounter complexity.}
\label{tab:cost_latency}
\begin{tabular}{lrr}
\toprule
Metric & Total & Per encounter \\
\midrule
\multicolumn{3}{l}{\textit{Latency}} \\
Processing time & -- & 76.8s \\
Throughput & -- & 46.9 enc/hr \\
\midrule
\multicolumn{3}{l}{\textit{Rounds executed}} \\
Total rounds & 2{,}087 & 1.41 \\
Terminal after Round 1 & 1{,}056 & 71.1\% \\
Terminal after Rounds 2--3 & 429 & 28.9\% \\
\midrule
\multicolumn{3}{l}{\textit{Token usage}} \\
Generation input tokens & 7.30M & 4{,}919 \\
Generation output tokens & 1.88M & 1{,}265 \\
Evaluation input tokens & 16.70M & 11{,}243 \\
Evaluation output tokens & 4.17M & 2{,}811 \\
\midrule
\multicolumn{3}{l}{\textit{API cost (USD)}} \\
Generation model & \$29.64 & \$0.0200 \\
Evaluation models & \$112.70 & \$0.0759 \\
\textbf{Total} & \textbf{\$142.33} & \textbf{\$0.0958} \\
\bottomrule
\end{tabular}
\end{table}
\section{Generator Prompt Rule Summary}
\label{app:prompt_rules}

The generator prompt enforces a conservative source-grounded operating point. The main
rules are: (1) use resolved clinical descriptions rather than raw codes; (2) include all
current-encounter structured evidence; (3) follow strict section structure; (4) use prior
encounters only as contextual support; (5) qualify all historical references; (6) do not infer
absent findings; (7) treat Subjective content as a conservative proxy when patient-reported
narrative is unavailable; (8) handle medication continuity conservatively and never infer
start dates; (9) keep Objective evidence-bearing only; (10) restrict Assessment to
current-encounter-grounded impressions; and (11) keep Plan grounded and non-empty.
Together, these rules prioritize traceability and hallucination control over unconstrained
narrative richness.

\section{Error Analysis of Round-1 Failures}
\label{app:error_analysis}

Table~\ref{tab:error_trajectory_app} reports the full recovery analysis for the 454
encounters that failed in Round~1. Categories are mutually exclusive and defined from
Round-1 threshold violations only. Recovery denotes eventual pass after up to two
additional refinement rounds.

\begin{table}[h]
\centering
\caption{Error trajectory of Round-1 failures. Confidence intervals are patient-clustered
95\% bootstrap intervals.}
\label{tab:error_trajectory_app}
\small
\begin{tabular}{lcccc}
\toprule
Failure mode at Round 1 & $N$ & Recovered & Recovery rate (95\% CI) & Discard rate \\
\midrule
Critical hallucination & 39 & 34 & 87.2\% [76.9, 97.1] & 12.8\% \\
Factuality failure & 158 & 124 & 78.5\% [70.9, 86.0] & 21.5\% \\
SFS failure & 116 & 71 & 61.2\% [51.1, 70.8] & 38.8\% \\
Completeness failure & 16 & 8 & 50.0\% [25.0, 75.0] & 50.0\% \\
Multiple simultaneous & 125 & 89 & 71.2\% [63.1, 79.1] & 28.8\% \\
\midrule
Total & 454 & 326 & 71.8\% [67.2, 76.4] & 28.2\% \\
\bottomrule
\end{tabular}
\end{table}

\section{Prompt Optimization on the Development Cohort}
\label{app:prompt_optimization}

We performed offline prompt optimization on a fixed 10-patient development cohort
containing 155 encounters. Each cycle followed three phases: diagnose recurring failures,
modify the base prompt, and validate on the same cohort. Table~\ref{tab:prompt_versions_app}
summarizes the version progression.

\begin{table}[h]
\centering
\caption{Prompt version comparison on the 10-patient development cohort.}
\label{tab:prompt_versions_app}
\small
\begin{tabular}{lrrrrrr}
\toprule
Version & Pass & Disc. & Rate & R0 Pass & R1 Rescue & R2 Rescue \\
\midrule
v1 & 137 & 18 & 88.4\% & 89 (57.4\%) & 37 (23.9\%) & 11 (7.1\%) \\
v2 & 130 & 25 & 83.9\% & 70 (45.2\%) & 45 (29.0\%) & 15 (9.7\%) \\
v3 & 142 & 13 & 91.6\% & 101 (65.2\%) & 27 (17.4\%) & 14 (9.0\%) \\
v4 & 146 & 9 & 94.2\% & 94 (60.6\%) & 37 (23.9\%) & 15 (9.7\%) \\
\bottomrule
\end{tabular}
\end{table}

\section{History-Selection Sensitivity Studies}
\label{app:history_selection}

We conducted two supporting analyses for longitudinal context. First, a pre-agentic
lookback-window study varied $k \in \{0,1,2,5\}$ using the earlier single-shot generator.
Completeness increased only modestly from $k=0$ to $k=5$, while available factuality
runs suggested possible carry-forward risk at longer windows; this motivated the default
choice of $k=2$. Second, a 30-patient relevance-based history pilot compared the default
recent-2 policy against selecting the two most similar prior encounters under a structured
overlap heuristic. The relevance-based selector underperformed the recent-2 baseline
(60\% vs.\ 87\% pass rate), so the final system retained the recent-history policy.

\begin{table}[h]
\centering
\caption{Pilot comparison of recent-history and relevance-based history selection.}
\label{tab:relevance_pilot_app}
\small
\begin{tabular}{lcc}
\toprule
Metric & Recent-2 & Relevant-2 \\
\midrule
Pass rate & 26/30 (87\%) & 18/30 (60\%) \\
Mean factual accuracy & 98.28\% & 85.67\% \\
Mean completeness & 97.92\% & 82.69\% \\
Mean structural fidelity & 74.72\% & 75.19\% \\
\bottomrule
\end{tabular}
\end{table}
\FloatBarrier
\section{Detailed Section Prediction Results}
\label{app:section_prediction}

Table~\ref{tab:bert_section_app} reports section-wise precision, recall, and F1 for the
paragraph-level section prediction task.

\begin{table}[h]
\centering
\caption{Section-wise performance on the MediSOAP test set.}
\label{tab:bert_section_app}
\small
\begin{tabular}{llccc}
\toprule
Model & Section & Precision & Recall & F1 \\
\midrule
MediSOAP-100 & Subjective & 1.000 & 0.228 & 0.371 \\
MediSOAP-100 & Objective & 0.553 & 0.976 & 0.706 \\
MediSOAP-100 & Assessment & 0.976 & 0.968 & 0.972 \\
MediSOAP-100 & Plan & 0.953 & 0.968 & 0.960 \\
\midrule
Synthetic & Subjective & 0.887 & 0.284 & 0.430 \\
Synthetic & Objective & 0.946 & 0.348 & 0.509 \\
Synthetic & Assessment & 0.485 & 0.976 & 0.648 \\
Synthetic & Plan & 0.769 & 1.000 & 0.870 \\
\midrule
Synthetic + MediSOAP-100 & Subjective & 0.939 & 0.988 & 0.963 \\
Synthetic + MediSOAP-100 & Objective & 0.983 & 0.916 & 0.948 \\
Synthetic + MediSOAP-100 & Assessment & 0.992 & 0.984 & 0.988 \\
Synthetic + MediSOAP-100 & Plan & 0.977 & 1.000 & 0.988 \\
\bottomrule
\end{tabular}
\end{table}
%%%%%%%%%%%%%%%%%%%%%%%%%%%%%%%%%%%%%%%%%%%%%%%%%%%%%%%%%%%%%%%%%%%%%%%%%%%%%%%
%%%%%%%%%%%%%%%%%%%%%%%%%%%%%%%%%%%%%%%%%%%%%%%%%%%%%%%%%%%%%%%%%%%%%%%%%%%%%%%

\end{document}